\documentclass[11pt]{article}
\usepackage{amsfonts}
\usepackage{amssymb}
\usepackage{graphicx}
\usepackage{color}
\usepackage{amsmath}
\usepackage{amsthm}
\usepackage{setspace}
\usepackage{fullpage}
\usepackage{natbib}

\usepackage{setspace}
\newcommand{\cX}{\mathcal{X}}

\newcommand{\defeq}{\mathrel{\mathop:}=}

\begin{document}
\renewcommand*{\thefootnote}{\fnsymbol{footnote}}

\title{\vspace{-1cm} Deep Learning in Finance}

\author{J. B. Heaton \footnote{Conjecture LLC, jb@conjecturellc.com} \and N. G. Polson \footnote {Booth School of Business,  
University of Chicago, ngp@chicagobooth.edu} \and J. H. Witte \footnote{Department of Mathematics, University College London, and Conjecture LLC, jhw@conjecturellc.com}}

\date{\small{February 2016}}

\maketitle


\begin{abstract}
\bigskip

\noindent We explore the use of deep learning hierarchical models for problems in financial prediction and classification. Financial prediction problems -- such as those presented in designing and pricing securities, constructing portfolios, and risk management -- often involve large data sets with complex data interactions that currently are difficult or impossible to specify in a full economic model.  Applying deep learning methods to these problems can produce more useful results than standard methods in finance. In particular, deep learning can detect and exploit interactions in the data that are, at least currently, invisible to any existing financial economic theory.\\

\end{abstract} \vspace{-.6cm}

\newpage
\singlespacing

\section{Introduction}

Financial prediction problems are of great practical and theoretical interest.  They are also quite daunting. Theory suggests that much information relevant to financial prediction problems may be spread throughout available economic and other data, an idea that also gains support from the many disparate data sources that different market participants watch for clues on future price movements.

Dealing with this variety of data sources is difficult. The collection of possibly relevant data is very large, while the importance of the data and the potentially complex non-linear interactions in the data are not well specified by financial economic theory. In practice, this results in a plethora of predictive models, many with little theoretical justification and subject to over-fitting and poor predictive out-of-sample performance.

What is needed is a method able to learn those complex features of the data inputs which lead to good predictions of the target output variables (such as an asset or portfolio return). 

In this paper, we introduce deep learning hierarchical decision models for problems in financial prediction and classification.  The deep learning predictor has a number of  advantages over traditional predictors, which include that 
\begin{itemize}
\item input data can be expanded to include all items of possible relevance to the prediction problem,
\item non-linearities and complex interactions among input data are accounted for, which can help increase in-sample fit versus traditional models,
\item over-fitting is more easily avoided. 
\end{itemize}

Our paper continues as follows. Section 2 introduces the deep learning framework. Section 3 presents three finance applications of the deep learning framework. Section 4 presents an example. Section 5 concludes.

A guiding principle throughout our paper is the construction of predictive models whose inputs are high-dimensional. See Breiman (2001) for a discussion that contrasts predictive algorithmic modelling with 
traditional statistical approaches.

\section{Deep Learning}\label{SecDL}

We begin by introducing the general theoretical deep learning framework as well as several specifications.

\subsection{Architecture}

Deep learning is a form of machine learning. Machine learning is using data to train a model and then using the trained model to make predictions from new data.  
The fundamental machine learning problem is to find a predictor of an output $Y$ given an input $X$. A learning machine is defined as
an input-output mapping $Y= F(X)$, where the input space is high-dimensional and we write
$$
Y = F(X ) \; \; {\rm where} \; \; X = ( X_1 , \ldots , X_p ),
$$
and a predictor is denoted by $ \hat{Y}(X) \defeq F(X)$.
The output $T$ can be continuous, discrete as in classification, or mixed.
For example, in a classification problem, we need to learn a mapping $ F :X \rightarrow Y$, where $Y \in \{ 1 , \ldots , K \} $ indexes categories.

 As a form of machine learning, deep learning trains a model on data to make predictions, but is distinguished by passing learned features of data through different \emph{layers} of abstraction. Raw data is entered at the bottom level, and the desired output is produced at the top level, the result of learning through many levels of transformed data. Deep learning is hierarchical in the sense that, in every layer, the algorithm extracts features into factors, and a deeper level's factors become the next level's features. 


Specifically, a deep learning architecture can be described as follows. Let $ f_1 , \ldots , f_L $ be given univariate activation functions for each of the $L$ layers. Activation functions are non-linear transformations of weighted data. A semi-affine activation rule is then defined by
$$
f_l^{W,b} \defeq f_l \left ( \sum_{j=1}^{N_l} W_{lj} X_j + b_l \right ) = f_l ( W_l X_l + b_l )\,,\quad 1\leq l\leq L,
$$
which implicitly needs the specification of the number of hidden units $N_l$. Our deep predictor, given the number of layers $L$, then becomes the composite map
\begin{equation*}
\hat{Y}(X) \defeq F(X) = \left ( f_1^{W_1,b_1} \circ \ldots \circ f_L^{W_L,b_L} \right ) ( X)\,.\label{DLComp}
\end{equation*}
Put simply, we model a high dimensional mapping, $F$, via the superposition of univariate semi-affine functions.
(Similar to a classic basis decomposition, the deep approach uses univariate activation functions to decompose a high dimensional $X$.)

We let $ Z^{(l)} $ denote the $l$-th layer, and so $ X = Z^{(0)}$.
The final output is the response $Y$, which can be numeric or categorical.
The explicit structure of a deep prediction rule is then
\begin{align*}
Z^{(1)} & = f^{(1)} \left ( W^{(0)} X + b^{(0)} \right ),\\
Z^{(2)} & = f^{(2)} \left ( W^{(1)} Z^{(1)} + b^{(1)} \right ),\\
\ldots  & \\
Z^{(L)} & = f^{(L)} \left ( W^{(L-1)} Z^{(L-1)} + b^{(L-1)} \right ),\\
\hat{Y} (X) & = W^{(L)} Z^{(L)} + b^{(L)}\,.
\end{align*}
Here, $W^{(l)} $ are weight matrices, and $b^{(l)} $ are threshold or activation levels.
Designing a good predictor depends crucially on the choice of univariate activation functions $ f^{(l)} $.

The $Z^{(l)}$ are hidden features (or factors) which the algorithm extracts. One particular feature is that the weight matrices  $ W_l  \in \mathbb{R}^{N_l \times N_{l-1}} $ are matrix valued.
This gives the predictor great flexibility to uncover non-linear features of the data -- particularly so in finance data, since the estimated hidden features
$ Z^{(l)} $ can represent portfolios of payouts. The choice of the dimension $ N_l $ is key, however, since
if a hidden unit (columns of $W_l$) is dropped  at layer $l$, then it eliminates all terms above it in the layered hierarchy. 


Put differently, the deep approach employs hierarchical predictors comprising of a series of $L$ non-linear transformations applied to $X$. 
Each of the $L$ transformations is referred to as a \emph{layer}, where the original input is $X$,
the output of the first transformation is the first layer, and so on, with the output $\hat{Y} $ as the $(L+1)$-th layer. 
We use $ l \in \{ 1 , \ldots , L \} $ to index the layers from $1$ to $L$, which are called \emph{hidden layers}. The number of layers $L$ represents the \emph{depth} of our architecture.

Commonly used activation functions are sigmoidal (e.g., $ 1 / (1 + \exp(-x)) $, $cosh (x)$, or $tanh(x)$), heaviside gate functions (e.g., $\mathbb{I}(x > 0 )$), or rectified linear units (ReLU) $\max\{x,0\}$.  ReLU's especially have been found to lend themselves well to rapid dimension reduction. A deep learning predictor is a data reduction scheme that avoids the curse of dimensionality through the
use of univariate activation functions. See Kolmorogov (1957), Lorenz (1976), Gallant and White (1988), Hornik et al. (1989), and Poggio and Girosi (1990)
for further discussion.

\subsection{Training a Deep Architecture}

Constructing a deep learner requires a number of steps. It is common to split the data-set into three subsets, namely training, validation, and testing.
The training set is used to adjust the weights of the network. The validation set is used to minimize the over-fitting and relates to the architecture design
(a.k.a. model selection). Finally, testing is used to confirm the actual predictive power of a learner. 

Once the activation functions, size, and depth of the learning routine have been chosen, we need to solve the training problem of finding $(\hat{W} , \hat{b}) $, 
where 
$$\hat{W}=(\hat{W}_0,\ldots,\hat{W}_L) \; \; {\rm and} \; \;  \hat{b}=(\hat{b}_0,\ldots,\hat{b}_L)$$ 
denote the learning parameters which we compute during training.
To do this, we need a training dataset $ D = \{ Y^{(i)} , X^{(i)} \}_{i=1}^T $ of input-output pairs and a loss function $\mathcal{L}(Y,\hat{Y})$ at the level of the output signal.
In its simplest form, we solve
\begin{equation}
{\rm arg\,min}_{W,b} \; \frac{1}{T} \sum_{i=1}^T \mathcal{L}( Y_i , \hat{Y}^{W,b}( X_i)) \,.\label{Training_Eq1}
\end{equation}
Often, the $L_2$-norm for a traditional least squares problem is chosen as error measure, and if we then minimize the loss function
\begin{equation*}
\mathcal{L}( Y_i , \hat{Y}( X_i)) = \|Y_i - \hat{Y}( X_i)\|^2_2\,,
\end{equation*}
our target function \eqref{Training_Eq1} becomes the mean-squared error (MSE) over the training dataset $ D = \{ Y^{(i)} , X^{(i)} \}_{i=1}^T $.

It is common to add a regularization penalty, denoted by $ \phi(W,b)$, to avoid over-fitting and to stabilize our predictive rule. We combine this with the loss function via a parameter $\lambda>0$, which gauges the overall level of regularization.
We then need to solve
\begin{equation}
{\rm arg\,min}_{W,b} \; \frac{1}{T} \sum_{i=1}^T \mathcal{L}( Y_i , \hat{Y}^{W,b} ( X_i)) + \lambda \phi(W,b)\,.\label{Training_Eq}
\end{equation}
The choice of the amount of regularization, $\lambda$, is a key parameter. This gauges the trade-off present in any statistical modelling that too little regularization will lead to over-fitting and poor out-of-sample performance.

In many cases, we will take a separable penalty, $ \phi(W,b) = \phi(W) + \phi(b) $. The most useful penalty is the ridge or $L^2$-norm, which can be viewed 
as a default choice, namely $$ \phi(W) = \Vert W \Vert^2_2 = \sum_{l=1}^T W_i^\top W_i . $$ Other norms include the lasso, which corresponds to an $L^1$-norm, and which can be
used to induce sparsity in the weights and/or off-sets.
The ridge norm is particularly useful when the amount of regularization, $\lambda$, has itself to be learned. This is due to the fact that 
there are many good predictive generalization results for ridge-type predictors.
When sparsity in the weights is paramount, it is common to use a lasso $L^1$-norm penalty.

\subsubsection{Probabilistic Interpretation}

In a traditional probabilistic setting, we could view the output $Y$ as a random variable generated by a probability 
model $p(Y| Y^{W,b}(X))$, where the conditioning is on the predictor $\hat{Y}(X)$. The corresponding loss
function is then 
$$\mathcal{L}(Y, \hat{Y} ) = - \log p( Y| Y^{ \hat{W} , \hat{b} } (X) ), $$ namely the negative log-likelihood.
For example, when predicting the probability of default, we have a multinomial logistic regression model which leads to a cross-entropy loss function. For multivariate normal models in particular (which includes many financial time series), the $L_2$-norm becomes a suitable error measure.

Probabilistically, the regularization term, $ \lambda \phi(W,b)$, can be viewed as a negative log-prior distribution over parameters, namely
\begin{align*}
 - \log p( \phi(W, b) ) & =  \lambda \phi(W,b),\\
 p( \phi(W, b) ) & =  C \exp ( - \lambda \phi(W,b)),
\end{align*}
where $C$ is a suitable normalization constant. This framework then provides a correspondence with Bayes learning. 
Our deep predictor is simply a regularized
maximum a posteriori (MAP) estimator. We can show this using Bayes rule as
\begin{align*}
 p( W, b | D ) & \propto  p( Y| Y^{W ,b } (X) ) p( W, b) \\
  & \propto  \exp \left ( - \log p( Y| Y^{W ,b } (X) ) - \log p( W, b) \right ),
\end{align*}
and the deep learning predictor satisfies
$$
\hat{Y} \defeq Y^{ \hat{W} , \hat{b} } (X) \; \; {\rm where} \; \; ( \hat{W} , \hat{b} ) \defeq 
{\rm arg \; min}_{W,b} \; \log p( W, b | D),
$$
and $$ - \log p( W, b | D ) = \sum_{i=1}^T \mathcal{L}( Y^{(i)} , Y^{W,b} (X^{(i)} ) ) + \lambda \phi( W, b )  $$ is the log-posterior distribution over parameters given the training data, $D = \{ Y^{(i)} , X^{(i)} \}_{i=1}^T$.

\subsubsection{Cross Validation}

Cross validation is a technique by which we split our training data into complementary subset to then conduct analysis and validation on different sets, aiming to reduce over-fitting and increase out-of-sample performance.

In particular, when training on time series, we may split our training data into disjoint time periods of identical length, which is particularly desirable in financial applications where reliable time consistent predictors are hard to come by and have to be trained and tested extensively.

Cross validation also provides a tool to decide  what levels of regularization lead to good generalization (i.e., prediction), which is the classic variance-bias trade-off. A key advantage of cross validation (over traditional statistical metrics such as $t$-ratios and $p$-values) is
that it also allows us to assess the size and depth of the hidden layers, that is, solve the model selection problem of choosing $L$ and $ N_l $ for $1 \leq l \leq L$ . This ability to pragmatically and seamlessly solve the model selection and estimation problems is one of the reasons for the current widespread use of machine learning methods.

\subsubsection{Back-propagation}

The common numerical approach for the solution of \eqref{Training_Eq} is a form of stochastic gradient descent, which adapted to a deep learning setting is usually called \emph{back-propagation}. One caveat of back-propagation in this context is the multi-modality of the system to be solved (and the resulting slow convergence properties), which is the main reason why deep learning methods heavily rely on the availability of large computational power. 

One of the advantages of using a deep network is that first-order derivative information is directly available. There are tensor libraries available that
directly calculate
$$ \nabla_{W,b} \mathcal{L}( Y_i , \hat{Y}^{W,b} ( X_i)) $$ using the chain rule across the training data-set. For ultra-large data-sets, we use mini-batches and
stochastic gradient descent (SGD) to perform this optimization, see LeCun et al. (2012). An active area of research is the use of this information within a Langevin MCMC algorithm that allows sampling from the full posterior distribution of the architecture. The deep learning model by its very design is highly multi-modal, and the parameters are high dimensional and in many cases unidentified in the traditional sense.
Traversing the objective function is the desired problem, and handling the multi-modal and slow convergence of traditional decent methods can be alleviated with proximal algorithms such as the
alternating method of multipliers (ADMM), as has been discussed in Polson et al. (2015 a, b). 

\subsection{Predictive Performance}

There are two key training problems that can be addressed using the predictive performance of an architecture.
\begin{enumerate}
\item[(i)] How much regularization to add to the loss function.
As indicated before, one approach is to use cross validation and to teach the algorithm to calibrate itself to a training data. An independent hold-out data set is kept separately to perform an 
out-of-sample measurement of the training success in a second step. As we vary the amount of regularization, we obtain a regularization path and choose the level
of regularization to optimize out-of-sample predictive loss. 
Another approach is to use
Stein's unbiased estimator of risk (SURE).
\item[(ii)]
A more challenging problem is to train the size and depth of each layer of the architecture, i.e., to determine
 $L$ and $N = ( N_1 , \ldots , N_L )$. This is known as the model selection problem. In the next subsection, we will describe a technique known as dropout,
which solves this problem. 
\end{enumerate}
Stein's unbiased estimator of risk (SURE) proceeds as follows. 
For a stable predictor, $\hat{Y}$, 
we can define the degrees of freedom of a predictor by  
$\text{\emph{df}} = \mathbb{E} \left(\sum_{i=1}^T \partial \hat{Y}_i/\partial{Y}_i \right ) $.
Then, given the scalability of our algorithm, the derivative 
$ \partial \hat{Y} / \partial Y $ is available using the chain
rule for the composition of the $L$ layers.

Now let the in-sample MSE be given by
$ \text{\emph{err}} = ||Y- \hat{Y}||^2_2 $ and, for a future observation $ Y^\star $, the out-of-sample predictive MSE is 
$$  \text{\emph{Err}} = \mathbb{E}_{ Y^\star} \left ( ||Y^\star - \hat{Y}||^2_2 \right ).$$ 
In expectation, we then have
$$
\mathbb{E} \left ( \text{\emph{Err}} \right ) = 
\mathbb{E} \left ( \text{\emph{err}} + 2 \text{Var} ( \hat{Y},Y ) \right).
$$  
The latter term can be written in terms of $ \text{\emph{df}} $ as a covariance.    
 Stein's unbiased risk estimate then becomes
$$
  \widehat{ \text{Err} } = ||Y- \hat{Y}||^2 
  + 2\sigma^2 \sum_{i=1}^{n} \frac{\partial \hat{Y}_i}{\partial Y_i}
  .
$$
As before, models with the best predictive MSE are favoured.

\begin{figure}[t]
  \centering
    \includegraphics[width=0.5\textwidth]{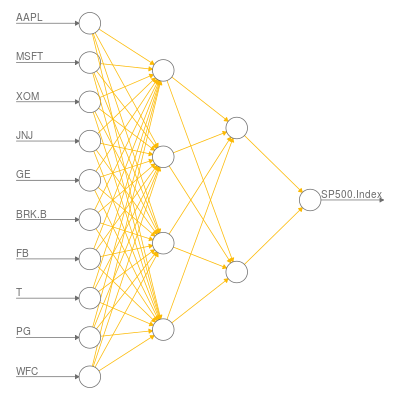}
  \caption{\footnotesize\emph{The hierarchical structure of several adaptive linear layers in a deep learning routine allows the extraction of non-linear features from the input data which can then be combined to a description of the desired target variable. This way, for dynamic in- and outputs, we obtain a deep feature policy (DFP) which for every combination of inputs tells us which corresponding action gives us the best approximation of the target variable. In the above picture, we see the setup for a DFP which through two hidden layers approximates the S\&P500 based on the ten largest companies in the index.
 \label{deepnet}
 }}
\end{figure}

\subsection{Dropout for Model Selection}

Dropout is a model selection technique. It is designed 
to avoid over-fitting in the training process, and does so by removing input dimensions in $X$ randomly with a given probability $p$. 
In a simple model with one hidden layer, we replace the network
\begin{align*}
Y_i^{(l)} & =  f ( Z_i^{(l)} ), \\
Z_i^{(l)} & = W_i^{(l)} X^{(l)}  + b_i^{(l)},
\end{align*}
with the dropout architecture
\begin{align*}
D_i^{(l)} & \sim \text{Ber} (p), \\
\tilde{Y}_i^{(l)} & = D^{(l)} \star X^{(l)}, \\
Y_i^{(l)} & =  f ( Z_i^{(l)} ), \\
Z_i^{(l)} & = W_i^{(l)} X^{(l)}  + b_i^{(l)}.
\end{align*}
In effect, this replaces the input $X$ by $ D \star X $, where $ \star$ denotes the element-wise product
and $D$ is a matrix of independent Bernoulli $\text{Ber}(p)$ distributed random variables. 

It is instructive to see how this affects the underlying loss function and optimization problem.
For example, suppose that we wish to minimise MSE,  $\mathcal{L}(Y,\hat{Y})=\|Y-\hat{Y}\|^2_2$, then, when marginalizing over the randomness, we have a new objective
$$
{\rm arg \; min}_W \; \mathbb{E}_{ D \sim {\rm Ber} (p) } \Vert Y - W ( D \star X ) \Vert^2_2\,,
$$
which is equivalent to
$$
{\rm arg \; min}_W \;  \Vert Y - p W X \Vert^2_2 + p(1-p) \Vert \Gamma W \Vert^2_2\,,
$$
where $ \Gamma = ( {\rm diag} ( X^\top X) )^{\frac{1}{2}} $. We can also interpret this last expression as a Bayesian ridge regression with a $g$-prior. Put simply, dropout reduces the likelihood of over-reliance on small sets of input data in training. See Hinton and Salakhutdinov (2006) and Srivastava et al. (2014). Lopes and West (2004) provide a fully Bayesian approach to factor selection. Dropout can be viewed as the optimization version of
the traditional spike-and-slab prior that has proven so popular in Bayesian model averaging.


Another application of dropout regularization is the choice of the number of hidden units in a layer. (This can be achieved if we drop units of the hidden rather than the input layer and then establish which probability $p$ gives the best results).
It is worth recalling though, as we have stated before, that one of the dimension reduction properties of a network structure
is that once a variable from a layer is dropped, all terms above it in the network also disappear. This is just the nature of a composite structure
for the deep predictor in \eqref{DLComp}.

We now turn to describing three widely used architecture designs that have become commonplace in applications of machine learning, namely auto-encoders, rectified neural networks (RNNs), and
long short term memory (LTSM) models. In Section \ref{smartindex}, we provide an application of auto-encoders to smart asset indexing problems.

\subsection{Auto-encoder}

An auto-encoder is a deep learning routine which trains the architecture to approximate $X$ by itself (i.e., $X=Y$) via a \emph{bottleneck} structure.
This means we select a model $F^{W,b} (X) $ which aims to concentrate the information required to recreate $X$. Put differently, an auto-encoder creates a more cost effective representation of $X$.

For example, under an $L^2$-loss function, we wish to find
$$
{\rm arg \; min}_{W,B} \; \Vert F^{W,b} (X) - X \Vert^2_2
$$
subject to a regulariziation penalty on the weights and offsets.

In an auto-encoder, for a training data set $ \{ X_1 , X_2 , \ldots \} $, we set the target values as $ Y_i = X_i$. A static auto-encoder with two linear layers, akin to a traditional factor model, can be written as a deep learner as
\begin{align*}
z^{(2)} & = W^{(1)} X + b^{(1)},\\
a^{(2)} & = f_2 ( z^{(2)} ),\\
z^{(3)} & = W^{(2)} a^{(2)} + b^{(2)},\\
Y = F^{W,b}(X) & = a^{(3)} = f_3 ( z^{(3)} ),
\end{align*}
where $ a^{(2)} , a^{(3)} $ are activation levels. It is common to set $ a^{(1)} = X $.
The goal is to learn the weight matrices $ W^{(1)} , W^{(2)} $.
If $ X \in \mathbb{R}^{N} $, then $ W^{(1)} \in \mathbb{R}^{M,N} $ and $ W^{(1)} \in \mathbb{R}^{N,M} $, 
where $ M \ll N $ provides the auto-encoding at a lower dimensional level.


If $ W_2$ is estimated from the structure of the training data matrix, then we have a traditional factor model, and the
$W_1$ matrix provides the factor loadings. (We note that PCA in particular falls into this category, as we have seen in \eqref{PCA_eq}.) If $W_2$ is estimated based on the pair $\hat{X}=\{Y,X\}=X$ (which means estimation of $W_2$ based on the structure of the training data matrix with the specific auto-encoder objective), then we have a sliced inverse regression model. If $W_1$ and $W_2$ are simultaneously estimated based on the training data $X$, then we have a two layer deep learning model. 

A dynamic one layer auto-encoder for a financial time series $(Y_t)$ can, for example, be written as a coupled system of the form
$$
Y_t = W_x X _t + W_y Y_{t-1} \; \; {\rm and} \; \; \left ( \begin{array}{c}
 X_t\\
 Y_{t-1} 
 \end{array}
 \right ) = W Y_t\,.
$$
We then need to learn the weight matrices $W_x$ and $W_y$. Here, the state equation encodes and 
the matrix $W$ decodes the $Y_t$ vector into its history $Y_{t-1}$ and the current state $X_t$.

The auto-encoder demonstrates nicely that in deep learning we do not have to model the variance-covariance matrix explicitly, as our model is already directly in
predictive form. (Given an estimated non-linear combination of deep learners, there is an implicit variance-covariance matrix, but that is not the driver of the method.) 

\begin{figure}[t]
  \centering
  \includegraphics[width=0.5\textwidth]{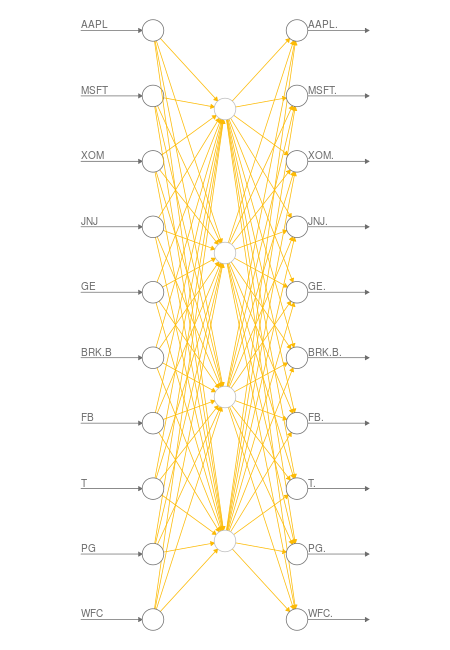}
  \caption{\footnotesize\emph{A deep auto-encoder depicted for the ten largest companies of the S\&P500. The hidden four unit layer of the deep auto-encoder compresses the aggregate information contained in all considered stocks and then produces a replication of every single input stock. If, instead of ten stocks, we compress all 500 stocks of the S\&P500 by sending them through the hidden bottleneck structure, we obtain a much more cost effective representation of the original index.}}
\label{autoensketch}
\end{figure}

\subsection{Long Short Term Memory Models (LSTMs)}

Traditional rectified neural nets (RNNs) can learn complex temporal dynamics via the set of deep recurrence equations
\begin{align*}
Z_t  & = f(W_{xz} X_t + W_{zz}  + b_x),\\
Y_t & =f(W_{hz} Z_t  +b_z),
\end{align*}
where $X_t$  is the input, $Z_t$ is the hidden layer with N hidden units, and $Y_t$ is the output at time $t$. For a length $T$ input sequence, the updates are computed sequentially.

Though RNNs have proven successful on tasks such as speech recognition and text generation (see Dean et al. 2012 and Lake et al. 2016), they have difficulty in learning long-term dynamics, due in part to the vanishing and exploding gradients that can result from propagating the gradients down through the many layers (corresponding to time) of the recurrent network. 

Long-short-term-memories (LSTMs) are a particular form of recurrent network which provide a solution by incorporating memory units. This allows the network to learn when to forget previous hidden states and when to update hidden states given new information. Models with hidden units with
varying connections within the memory unit have been proposed in the literature with great empirical success.
 Specifically, in addition to a hidden unit $Z_t$, LSTMs include an input gate,  a forget gate, an input modulation gate, and a memory cell. 
 The memory cell unit combines the previous memory cell unit which is modulated by the forget and input modulation gate together with the previous hidden state, modulated by the input gate.
 These additional cells enable an LSTM architecture to learn extremely complex long-term temporal dynamics that a vanilla RNN is not capable of. Additional depth can be added to LSTMs by stacking them on top of each other, using the hidden state of the LSTM as the input to the next layer.
 
An architecture for an LSTM model might be
\begin{align*}
F_t &= \sigma(W_f^T[Z_{t-1},X_t] + b_f),\\
I_t &= \sigma(W_i^T[Z_{t-1},X_t] + b_i),\\
\bar{C}_t &= \tanh(W_c^T[Z_{t-1},X_t] + b_c),\\
C_t &= F_t \otimes C_{t-1} + I_t \otimes \bar{C}_t,\\
Z_t &= O_t \otimes \tanh(C_t).
\end{align*}
The key addition, compared to an RNN, is the hidden state $C_t$, the information is added or removed from the memory state via layers defined via a sigmoid function $\sigma(x) = (1+e^{-x})^{-1}$ and point-wise multiplication $\otimes$. The first gate $ F_t \otimes C_{t-1}$, called the forget gate, allows to throw away some data from the previous cell state. The next gate, $ I_t \otimes \bar{C}_t$, called the input  gate, decides  which values will be updated. Then the new cell state is a sum of the previous cell state, passed through the forgot gate selected components of the $[Z_{t-1},X_t]$ vector. This provides a mechanism for dropping irrelevant information from the past and adding relevant information from the current time step. Finally, the output layer, $O_t \otimes \tanh(C_t)$, returns \emph{tanh} applied to the hidden state with some of the entries removed. 

An LSTM model might potentially improve predictors by utilizing data from the past by memorizing volatility patterns from previous periods. The LSTM model  allows to automate the identification of the temporal relations in the data, at the cost of larger sets of parameters to be trained.  

There are numerous finance applications of LTSM models. They provide a new class
of volatility models that are capable of capturing long-memory effects in the underlying structure of asset return movements.

\begin{figure}[t]
  \centering
    \includegraphics[width=0.9\textwidth]{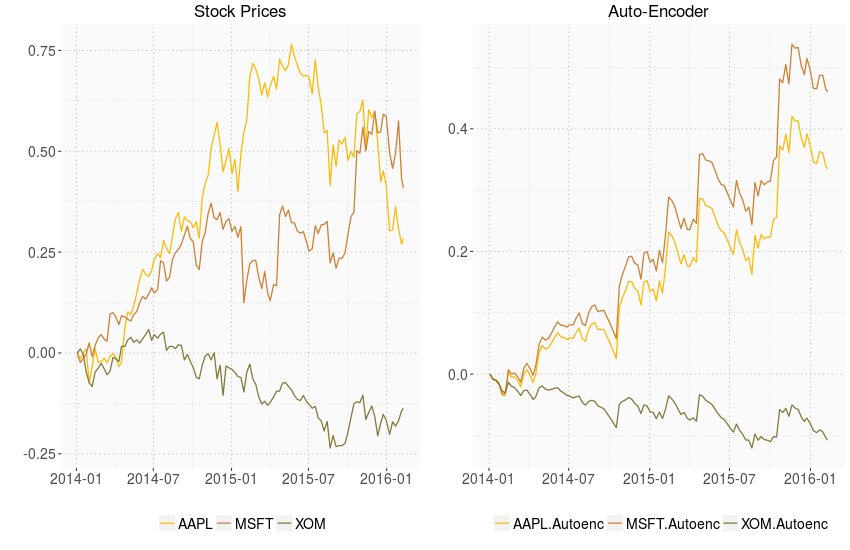}
  \caption{\footnotesize\emph{For the time period 2014/15, we see AAPL, MSFT, and XOM stock prices before (left) and after auto-encoding (right).}}
\label{SP500Autoenc_Example}
\end{figure}

\section{Finance Applications}

We now come to discuss deep learning specifically in the context of finance.
For areas of finance applications, see Fama and French (1992, 2008), Engle (1982), Campbell, Lo, and MacKinley (1997), Singleton (2006), and Daniel and Titman (2006).  
Hutchison, Lo, and Poggio (1994) provide a shallow learner for option pricing.

\subsection{Deep Factor Models versus Shallow Factor Models}


Almost all shallow data reduction techniques can be viewed as consisting of a low dimensional auxiliary
variable $Z$ and a prediction rule specified by a
composition of functions 
\begin{align*}
  \hat{Y} &= f_1^{W_1,b_1} (f_2( W_2X +b_2)\big) 
  \\
  &= f_1^{W_1,b_1}(Z),\,\text{ where $Z:=f_2(W_2X+b_2)$. }
\end{align*}
In this formulation, we also recognize the previously introduced deep learning structure \eqref{DLComp}.
The problem of high dimensional data reduction in general is to find the $Z$-variable and to estimate the layer functions $(f_1, f_2)$ correctly. In the layers, we want to uncover the low-dimensional $Z$-structure in a way that does not disregard information about predicting the output $Y$.

Principal component analysis (PCA), reduced rank regression (RRR), linear discriminant analysis (LDA), project pursuit regression (PPR), and logistic regression are all shallow learners. 
See Wold (1956), Diaconis and Shahshahani (1984), Ripley (1996), Cook (2007), and Hastie et al. (2009) for further discussion.

For example, PCA reduces $X$ to $f_2(X)$ using a
singular value decomposition of the form
\begin{equation}
Z =  f_2(X) = W^\top X + b\,,\label{PCA_eq}
\end{equation}
where the columns of the weight matrix $W$ form an orthogonal basis for directions of greatest variance (which is in effect an eigenvector problem). Similarly, for the case of $X=(x_1,\ldots,x_p)$, PPR reduces $X$ to $f_2(X)$ by setting
$$
Z=f_2(X) = \sum^{N_1}_{i=1}f_i(W_{i1}X_1 + \ldots + W_{ip}X_p)\,.
$$
As stated before, these types of dimension reduction are
independent of $y$ and can easily discard information that is valuable for
predicting the desired output. Sliced inverse regression (SIR) overcomes this drawback somewhat by
estimating the layer function $f_2$ using data on both, $Y$ and $X$, but still operates independently of $f_1$.

Deep learning overcomes many classic drawbacks by \textit{jointly}
estimating $f_1$ and $f_2$ based on the full training data $\hat{X}=\{Y_i,X_i\}^T_{i=1}$, using information on $Y$ and $X$ as well as their relationships, and by using $L>2$ layers.
If we choose to use non-linear layers, we can view a deep learning routine as a hierarchical non-linear factor model, or, more specifically, as a generalized linear model (GLM) with recursively defined non-linear link functions.



Stock and Watson have used a similar approach when forecasting a single time series on inflation based on a large numbers of predictors. Another obvious application is cost effective indexing replication, where we are trying to create a small sub-portfolio which dynamics similar to the main index (see also Section \ref{smartindex}).

\subsection{Default Probabilities}

Another area of great application of deep learning is credit risk analysis. The goal of a deep learning model is a feature representation of a high dimensional input space.
For example, in image processing, one can think of the layers as first representing objects, then object parts (faces), then edges, and finally pixels.
A similar \emph{feature map} can be found for the credit-worthiness of companies. We can combine financial asset return data with text data (earnings calls) and accounting data (book values, etc.)
to obtain an \emph{image} of the health of a firm.

Specifically, suppose that our observations $Y_i$ represents a multi-class $1$-of-$K$
indicator vector. We equate classes via $Y=k$ for $1 \leq k \leq K$.
For example, the $K$ classes might correspond to bond ratings. At the extreme, we might have an indicator $ y_i \in \{0,1\}$ which indicates
bankrupt or not.  

We need to model the probability of default.
Suppose that $Y_i\in\{-1,1\}$ is categorical. Given the output $X$, it is common to model the probability of default via a soft-max (or logic) activation function
$$
p(Y_i\mid W,b,X) = \frac{1}{1+e^{\hat{Y}^{W,b} (X)}}\,,
$$
where $\hat{Y}^{W,b}(X) =W_1f_1(\ldots + b_1)$. We define $\hat{p}(X_i) = \text{arg max }_{W,b} \;  p(Y_i\mid W,b,X_i)$ as the maximum probability estimator. 

Given a multinomial likelihood $p(Y,\hat{Y})$, this then leads to a cross-entropy loss function 
$$\mathcal{L}(Y,\hat{Y}) = -\log p(Y,\hat{Y}) = -Y \log \hat{p}(X).$$ Alternatively, in its natural parameter space, we have the log-odds as a deep predictor $\hat{Y}_i = \log (p_i/(1-p_i)$.
We have a multi-class predictor
  $$
  p(Y_i=k|\hat{Y}_i^{\hat{W}, \hat{b}} (X)) = \sigma_k(W_1 Z_1),  
  $$
  where $\sigma(x) = 1 / 1 + e^x$. The negative log-likelihood is given by
  \begin{equation*}
    \mathcal{L}(Y_i, \hat{Y}_i) = - \log p(Y_i|\hat{Y}_i) 
    = -\log \prod_{k=1}^{K} (\hat{Y}_{i,k})^{Y_{i,k}} = -\sum_{k=1}^{K} Y_{i,k} \log \hat{Y}_{i,k}.
  \end{equation*}
Therefore, minimizing the cross-entropy cost functional is equivalent to a
  multi-class logistic likelihood function
The gain from a deep learning approach is our ability to include the \emph{kitchen-sink} into the input space $X$.
Feature extraction is the main output from a deep learner and these non-linear contrast of input variables provide summaries of the 
tendency for firms to default. 


\subsection{Event Studies}
Let $y \in \mathcal{S}$ denote an observed output variable, where 
$\mathcal{S} = \mathbb{R}^N$ for regression and $\mathcal{S} = \{1, \ldots, K\}$ for
classification. We have a  high dimensional input/covariate variable given by
$\cX = \{X_t\}_{t=1}^T$ and $X_t \in \mathbb{R}^{N \times M} $.  

We can build a deep learner for event study analysis as follows.
Given a series of input event embeddings $ X = ( X_1 , \ldots , X_n ) $, we can use a weight matrix $ W_1 \in \mathbb{R}^l $ to extract the $l$-possible events.
For example, $l=4$ for earnings announcements during the year, and $n=252$ for number of trading days. We now construct
a hidden factor
$$
Z_j = W_1^\top X_{j-l+1},
$$
so we can measure the effect of the $l$ previous events on today's return.

We might use a max-pooling activation function if we think that the effect is based solely on the largest value (i.e., $\max Z_j$), in which case the model ignores all other and focuses on the largest.


\begin{figure}[p]
  \centering
    \includegraphics[width=0.9\textwidth]{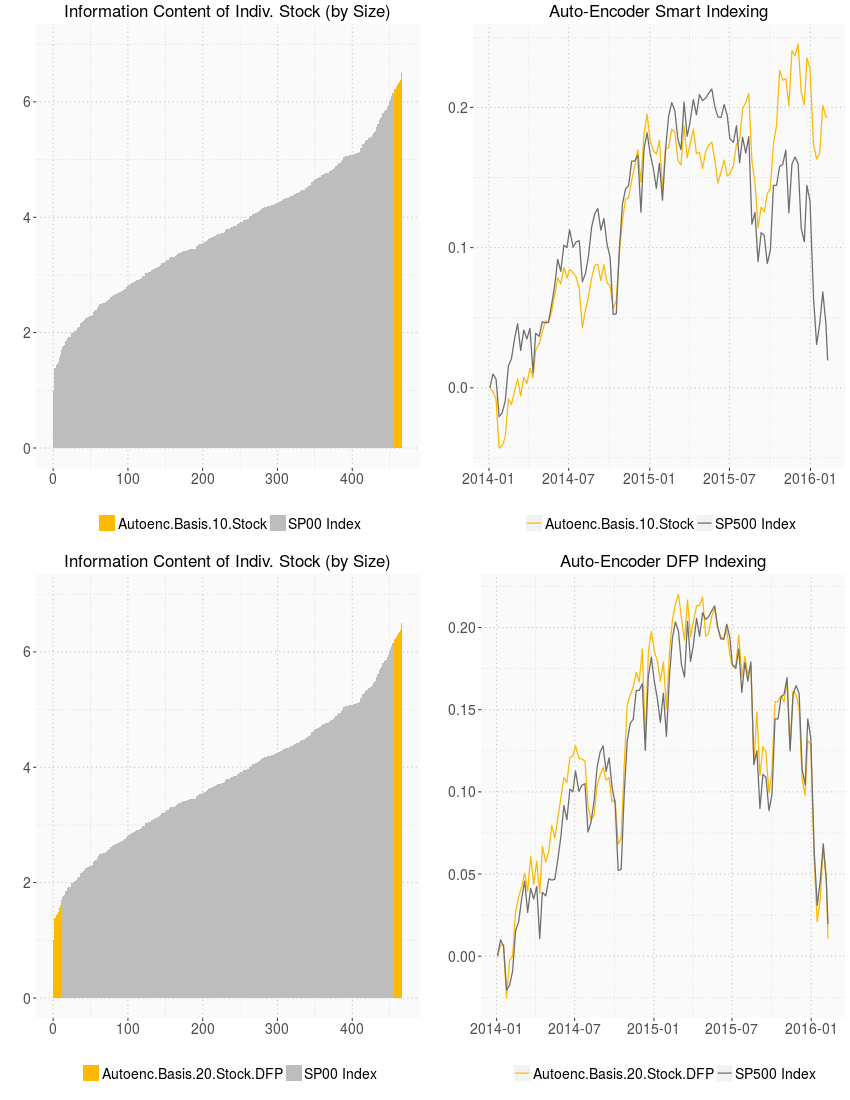}
  \caption{\footnotesize\emph{A deep auto-encoder compresses the the stocks of the S\&P500. Above, we rank all stocks by their proximity to the auto-encoded information and create an equally weighted portfolio from the auto-encoder basis of the 10 leading stocks. Below, we use the leading ten (the most \emph{communal} stocks) and the bottom ten (most \emph{individualistic}) stocks to create an auto-encoder basis on which we train a deep feature policy (DFP) for the approximation of the S\&P500.}}
\label{smartindexbig}
\end{figure}

\section{Example: Smart Indexing}\label{smartindex}

When aiming to replicate (or approximate) a stock index through a subset of stocks, there are two conceptual approaches we can choose from.
\begin{itemize}
\item[{\bf (i)}] Identify a small group of stocks which historically have given a performance very similar to that of the observed index.
\item [{\bf (ii)}] Identify a small group of stocks which historically have represented an over-proportionally large part of the total aggregate information of all the stocks the index comprises of. 
\end{itemize}

While, on the face of things, (i) and (ii) may appear very similar, they characterize, in fact, very different methodologies. 

Many classic approaches to index replication are essentially rooted in linear-regression, which is part of group (i). Frequently, by trial and error, we are trying to find a small subset of stocks which in-sample gives a reasonable linear approximation of the considered index.

The deep learning version of (i) allows translating the input data through a hierarchical sequence of adaptive linear layers into a desired output, which means that, in training, even non-linear relationships can be readily identified. Since every hidden layer provides a new interpretation of the input features, we refer to the resulting strategy for approximation (or prediction) as a deep feature policy (DFP), an example of which is given in Figure \ref{deepnet}.

The availability of tailored non-linear relationships in deep learning makes the conventional objective of (i), namely good in-sample approximation, an easily achieved triviality, and takes the focus straight to training for out-of-sample performance (which brings us back to cross-validation and dropout, see Section \ref{SecDL}).

Another weakness of (i) is that it fits the finished (and through aggregation diluted) product. A deep auto-encoder avoids this problem by directly (rather than indirectly) approximating the aggregate information contained in the considered family of index stocks. In Figure \ref{autoensketch}, a deep auto-encoder for a small set of ten stocks is depicted. In Figure \ref{SP500Autoenc_Example}, we see the stock paths before and after compression.\footnote{We use an auto-encoder with one hidden layer of 4 units and a sparsity constraint of $\rho=0.01$ (to avoid training the identity function).}

The bottleneck structure of an auto-encoder creates a compressed set of information from which all stocks are re-created (through linear and non-linear relationships). Thus, for indexing, the stocks which are closest to the compressed core of the index can be interpreted as a non-linear basis of the aggregate information of the considered family of stocks.

In Figure \ref{smartindexbig} at the top, we auto-encoded all stocks of the S\&P500 over the period 2014/15. We then ranked the stocks by how close they were to their own auto-encoded version; the closer, the higher the \emph{communal} information content of a stock. On the top right, we see the approximation of the S\&P500 obtained by simply investing in the ten stocks with the highest communal information content. 

We notice that, while the ten stock auto-encoder basis is reasonable, the approximation is a little off, particularly in the last seven months of the training period. It is instructive to observe how in-sample this deviation can easily be avoided by using a DFP index approximation.

In Figure \ref{smartindexbig} at the bottom, we combined the two sets of ten stocks with the highest and lowest communal information, respectively, and then trained a deep learning routine over the same period 2014/15 to approximate the S\&P500 index based on this expanded basis of twenty stocks.\footnote{We use a deep neural net with (4,2) hidden layers.} For every combination of inputs from the selected twenty stocks, the DFP gives us, based on the hierarchical composition of non-linear features extracted from the input data, an optimized action for the approximation of the desired index.

Given sufficient diversity of the input data, a DFP can often be trained to approximate the target data to almost arbitrarily accuracy, an improvement we now notice for the last six months of the training period in the bottom right chart in Figure \ref{smartindexbig}.

In short, many classic models have had to focus on the wrong thing, namely in-sample approximation quality, due to their shortcomings in that area, while deep learning naturally addresses out-of-sample performance as optimization target.

\begin{figure}[t]
  \centering
    \includegraphics[width=0.9\textwidth]{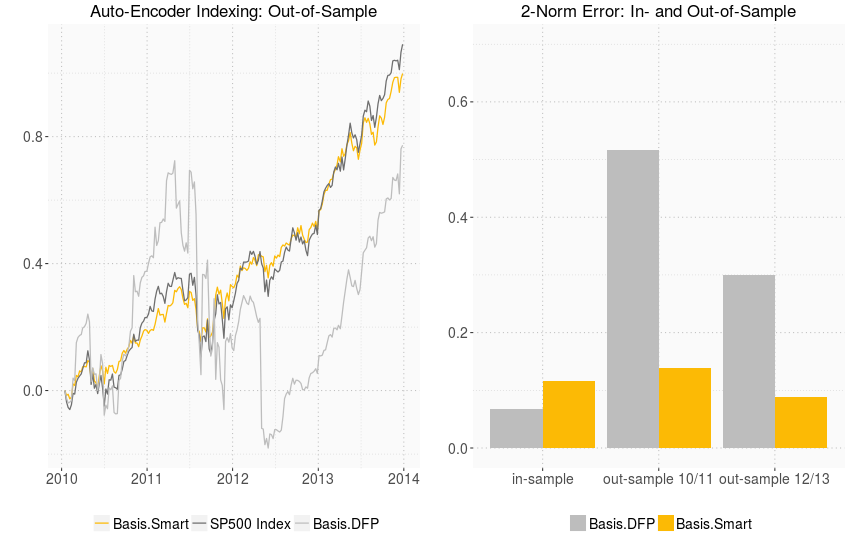}
  \caption{\footnotesize\emph{Having trained a 20 stock DFP auto-encoder basis as well as a simpler 10 stock deep auto-encoder basis for the S\&P500 on the two years 2014/15, we now consider the out-of-sample performance of the two approximations for the periods 2010/11 and 2012/13. We observe that the superior fitting qualities of the full DFP basis in-sample are out-weighed easily by the superior out-of-sample consistency of the generic deep auto-encoder basis.}}
\label{SmartIndex_OutSample}
\end{figure}

In Figure \ref{SmartIndex_OutSample}, we apply our two example index trackers to the period 2010/14 as an out-of-sample test. We notice how the previously superior DFP approximation is unreliable, while the simple auto-encoder basis (made up of ten stocks rich in communal information) provides a consistent index replication. We conclude that, for index replication, auto-encoding as suggested by (ii) seems to be the more robust approach, and that the superior learning abilities of a DFP require careful handling in training to achieve the desired result.  

%
%
%
%
%
%
%

\section{Conclusion}

Deep learning presents a general framework for using large data sets to optimize predictive performance. As such, deep learning frameworks are well-suited to many problems -- both practical and theoretical -- in finance. This paper presents deep learning hierarchical decision models for problems in financial prediction and classification. As we have demonstrated, deep learning has the potential to improve -- sometimes dramatically -- on predictive performance in conventional applications. Our example on smart indexing in Section \ref{smartindex} presents just one way to implement deep learning models in finance. Many other applications remain for development. 

At the same time, deep learning is likely to present significant challenges to current thinking in finance, including, most notably, the concept of market efficiency. Because it can model complex non-linearities in the data, deep learning may be able to \emph{price} assets to within arbitrarily small pricing errors. Will this imply that markets are informationally efficient, or will new tests of market efficiency be necessary? Overall, it is unlikely that any theoretical models built from existing axiomatic foundations will be able to compete with the predictive performance of deep learning models. What this means for the future of financial economics remains to be seen.

In the meantime, deep learning models are likely to exert greater and greater influence in the practice of finance, particularly where prediction is paramount.

\bigskip
\vspace{0.5cm}


{\Large\bf References}

[1] L. Breiman: \emph{Statistical modeling: the two cultures (with comments and a rejoinder
by the author)}. {\bf Statistical Science}, Vol. 16(3), pp. 199-231, 2001.

[2] J. Y. Campbell, A. W. Lo and A. C. MacKinley: \emph{The econonmetrics of financial markets}. {\bf Princeton University Press}, 1997.

[3] R. D. Cook: \emph{Fisher lecture: dimension reduction in regression}, {\bf Statistical
Science}, pp. 1-26, 2007.

[4] K. Daniel and S. Titman: \emph{Market reactions to tangible and intangible information}, {\bf Journal of Finance}, Vol. 61, 1605-1643, 2006.

[5] J. Dean, G. Corrado, R. Monga, et al.: \emph{Large scale
distributed deep networks}, {\bf Advances in Neural Information Processing Systems},
pp. 1223-1231, 2012.

[6] P. Diaconis and M. Shahshahani: \emph{On non-linear functions of linear combinations}, {\bf SIAM 
Journal on Scientific and Statistical Computing}, Vol. 5(1), pp. 175-191, 1984.

[7] R. Engle: \emph{Autoregressive conditional heteroscedasticity with estimates of the variance of United Kingdom inflation}, {\bf Econometrika}, 50(4), 987-1007, 1982.

[8] E. F. Fama and K. R. French: \emph{The cross-section of expected stock returns}, {\bf Journal of Finance}, Vol. 47, pp. 427-465, 1992.

[9] E. F. Fama and K. R. French: \emph{Dissecting anomalies}, {\bf Journal of Finance}, Vol. 53(4), pp. 1653-1678, 2008.

[10] A. R. Gallant and H. White: \emph{There exists a neural network that does not make avoidable mistakes}, {\bf IEEE International Conference on Neural Networks}, Vol. 1, pp. 657-664, 1988. 

[11] T. Hastie, R. Tibshirani, and J. Friedman: \emph{The elements of statistical learning}, Vol 2, 2009.

[12] G. E. Hinton and R. R. Salakhutdinov: \emph{Reducing the dimensionality of
data with neural networks}, {\bf Science}, Vol. 313(5786), pp. 504-507, 2006.

[13] K. Hornik, M. Stinchcombe, and H. White: \emph{Multilayer feedforward networks are universal 
approximators}, {\bf Neural networks}, Vol. 2(5), pp. 359-366, 1989.

[14] J. M. Hutchinson, A. W. Lo, and T. Poggio: \emph{A Nonparametric approach to pricing and hedging derivative securities
via learning networks}, {\bf Journal of Finance}, Vol. 48(3), pp. 851-889, 1994.

[15] A. Kolmogorov: \emph{The representation of continuous functions of many variables
by superposition of continuous functions of one variable and addition}, {\bf Dokl. Akad. Nauk SSSR}, Vol. 114, pp. 953-956, 1957.

[16] B. M. Lake, R. Salakhutdinov, and J. B. Tenenbaum: \emph{Human-level concept learning through probabilistic program induction}. {\bf Science}, Vol. 3560, pp. 1332-1338, 2015

[17] Y. A. LeCun, L. Bottou, G. B. Orr, and K.-R. Muller: \emph{Efficient
backprop}, {\bf Neural networks: Tricks of the trade}, pp. 9–48, 2012.

[18] H. F. Lopes and M. West: \emph{Bayesian Model Assessment in Factor Analysis}. {\bf Statistica Sinica}, 14, pp. 41-67, 2004.

[19] G. G. Lorentz: \emph{The 13th problem of Hilbert}, {\bf Proceedings of Symposia in Pure Mathematics}, American Mathematical Society,
Vol. 28, pp. 419-430, 1976.

[20] T. Poggio and F. Girosi: \emph{Networks for approximation and learning}, {\bf Proceedings of the IEEE}, Vol. 78(9), pp. 1481-1497, 1990.

[21] N. G. Polson, J. G. Scott, and B. T. Willard: \emph{Proximal algorithms
in statistics and machine learning}, {\bf Statistical Science}, 30, 559-581, 2015.

[22] N. G. Polson, B. T. Willard, and M. Heidari: \emph{A statistical theory for Deep Learning}, 2015.

[23] B. D. Ripley: \emph{Pattern recognition and neural networks}. {\bf Cambridge University
Press}, 1996.

[24] K. J. Singelton: \emph{Empirical Dynamic Asset Pricing}. {\bf Princeton Univertsity Press}, 2006.

[25] Srivastava et al.: \emph{Dropout: a simple way to prevent neural networks from overfitting}, {\bf 
Journal of Machine Learning Research}, 15, 1929-1958, 2014.

[26] H. Wold: \emph{Causal inference from observational data: a review of end and
means}, {\bf Journal of the Royal Statistical Society}, Series A (General), pages 28-61, 1956.

\end{document}